\documentclass[twocolumn,a4paper,14pt]{IEEEtran}
\pdfoutput=1
\usepackage{amsmath} 
\usepackage{amssymb}
\usepackage{graphicx}
\usepackage{subfigure} 
\usepackage{float} 
\usepackage[numbers]{natbib}
\usepackage{tcolorbox}
\usepackage{rotating}
\usepackage{url}

\newcommand{\mywidth}{1.5in}
\newcommand{\mytable}{sidewaystable}

\newcommand{\etal}{{et al}.~}
\newcommand{\dotp}[2]{\ensuremath{\langle #1 , #2 \rangle}}
\newcommand{\medto}{\ensuremath{\operatorname{median}_2}}
\newcommand{\Medto}{\ensuremath{\operatorname{Median}_2}}
\newcommand{\fone}{\ensuremath{\operatorname{F}_1}}
\begin{document} 
\title{The Mean and Median Criterion for Automatic Kernel Bandwidth Selection for Support Vector Data Description}

\author{\
\IEEEauthorblockN{
Arin Chaudhuri\IEEEauthorrefmark{1}, 
Deovrat Kakde\IEEEauthorrefmark{2},
Carol Sadek\IEEEauthorrefmark{3},
Laura Gonzalez\IEEEauthorrefmark{4},
Seunghyun Kong\IEEEauthorrefmark{5}
}

\IEEEauthorblockA{
Advanced Analytics Divison,
SAS Institute Inc.\\
Cary, NC, USA\\
Email: 
\IEEEauthorrefmark{1}Arin.Chaudhuri@sas.com,
\IEEEauthorrefmark{2}Dev.Kakde@sas.com,
\IEEEauthorrefmark{3}sadekcw@live.unc.edu,
\IEEEauthorrefmark{4}Laura.Gonzalez@sas.com,
\IEEEauthorrefmark{5}Seunghyun.Kong@sas.com,
}
}

\maketitle
\begin{abstract} 
Support vector data description (SVDD) is a popular
technique for detecting anomalies. The SVDD classifier
partitions the whole space into an \emph{inlier} region,
which consists of the region \emph{near} the training data,
and an \emph{outlier} region, which consists of points
\emph{away} from the training data. The computation of
the SVDD classifier requires a kernel function, and the
Gaussian kernel is a common choice for the kernel function.
The Gaussian kernel has a bandwidth parameter, whose value
is important for good results. A small bandwidth leads to
overfitting, and the resulting SVDD classifier overestimates
the number of anomalies. A large bandwidth leads to
underfitting, and the classifier fails to detect many anomalies.
In this paper we present a new automatic, unsupervised
method for selecting the Gaussian kernel bandwidth.
The selected value can be computed quickly, and it is
competitive with existing bandwidth selection methods.
\end{abstract}

\section{Introduction}
\addtocontents{toc}{\protect\setcounter{tocdepth}{2}}
\label{intro}
Support vector data description (SVDD) is a machine learning
technique that is used for single-class classification
and anomaly detection. First introduced by Tax and Duin
\cite{tax2004support} SVDD's mathematical formulation is
almost identical to the one-class variant of support vector
machines : one-class support sector machines (OCSVM), which 
is attributed to
Sch{\"ol}kopf \etal \cite{scholkopf2000support}. The use
of SVDD is popular in domains where the majority of data
belongs to a single class and it is not possible to make
any distributional assumptions. For example, SVDD is useful
for analyzing sensor readings from reliable equipment where
almost all the readings describe the equipment's normal
state of operation.

Like other one class classifiers SVDD provides a geometric description
of the observed data. The SVDD classifier assigns a \emph{distance} to each point in
the domain space; which measures the separation of that point from the
training data. During scoring any observation found to be at a \emph{large}
distance from the training data might be an anomaly, and the user might choose
to generate an alert.

Several researchers have proposed using SVDD for multivariate process
control \cite{sukchotrat2009one,camci2008general}. Other applications of SVDD involve machine condition monitoring \cite{widodo2007support,ypma1999robust} and image classification \cite{sanchez2007one}.

\subsection{Mathematical Formulation}
In this subsection we describe the mathematical formulation of SVDD, the description is based on
\cite{tax2004support}.\mbox{}\\
{\bf Normal Data Description:}\\
The SVDD model for normal data description builds a hypersphere that contains
most of the data within a small radius. Given observations $x_1,\dots,x_n$, 
we need to solve the follwing optimization problem to obtain the SVDD data description.

\ \\
{\bf Primal Form:}\\
Objective:
\begin{equation}
\mathrm{min}~ R^{2} + C\sum_{i=1}^{n}\xi_{i}, 
\end{equation}
subject to: 
\begin{align}
 \|x_{i}-a\|^2 \leq R^{2} + \xi_{i}, \forall i=1,\dots,n,\\
 \xi_{i}\geq 0, \forall i=1,...n.
\end{align}
where:\\
$x_{i} \in {\mathbb{R}}^{m}, i=1,\dots,n  $ represent the training data,\\
$R$ is the radius and represents the decision variable,\\
$\xi_{i}$ is the slack for each variable,\\
$a$ is the center (a decision variable), \\
$C=\frac{1}{nf}$ is the penalty constant that controls the trade-off between the volume and the errors, and\\
$f$ is the expected outlier fraction.\\ \ \\
{\bf Dual Form:}\\
The dual formulation is obtained using Lagrange multipliers.\\ 
Objective:
\begin{equation} 
    \textrm{max}~ \sum_{i=1}^{n}\alpha _{i}\dotp{x_{i}}{x_{i}} - 
\sum_{i,j}^{ }\alpha _{i}\alpha _{j}\dotp{x_{i}}{x_{j}} ,
\end{equation}
subject to:
\begin{align}
\sum_{i=1}^{n}\alpha _{i} = 1\\
0 \leq  \alpha_{i}\leq C,\forall i=1,\dots,n
\end{align}
where $\alpha_{i}\in \mathbb{R}$ are the Lagrange constants, and $C=\frac{1}{nf}$ is the penalty constant.\\ \ \\
{\bf Duality Information:}\\
The position of observation $x_i$ is connected to the optimal $\alpha_i$, the radius of the
sphere $R$, and the center of the sphere $a$ in the
following manner: 

Center position: 
\begin{equation} \sum_{i=1}^{n}\alpha _{i}x_{i}=a \end{equation}

Inside position: 
\begin{equation} \left \| x_{i}-a \right \| < R \iff \alpha _{i}=0\end{equation}

Boundary position: 
\begin{equation} \left \| x_{i}-a \right \| = R \iff 0< \alpha _{i}< C\end{equation}

Outside position: 
\begin{equation}\left \| x_{i}-a \right \| > R \iff \alpha _{i}= C\end{equation}

Any $x_i$ for which the corresponding $\alpha_i > 0$ is known as a support vector.

Let $SV_{<~C}$ denote the set $\{ x_j : 0 < \alpha_j < C \}$
then the radius of the hypersphere is calculated, for any
$x_k \in SV_{<~C}$, as follows:
\begin{equation}   
R^{2}=\dotp{x_{k}}{x_{k}}-2\sum_{i}^{ }\alpha _{i}\dotp{x_{i}}{x_{k}}+\sum_{i,j}^{ }\alpha _{i}\alpha _{j}\dotp{x_{i}}{x_{j}}.
 \end{equation}
The value of $R^2$ does not depend on the choice of $x_k \in SV_{<~C}$.
\mbox{}\\
{\bf Scoring:}

For any point $z$,  the distance $ \textrm{dist}^{2}(z) $ is calculated as follows: 
\begin{equation}  \textrm{dist}^{2}(z)= \langle z, z \rangle - 2\sum_{i}^{ }\alpha _{i}\langle x_{i}, z \rangle +\sum_{i,j}^{}\alpha_{i}\alpha_{j}\langle x_{i},x_{j}\rangle \end{equation}
Points whose $\textrm{dist}^{2}(z) > R^{2} $ are designated as outliers.

The spherical data boundary can include a significant
amount of space that has a sparse distribution of training
observations. Using this model to score can lead to a 
lot of false positives. Hence, instead of a
spherical shape, a compact bounded outline around the data
is often desired. Such an outline should approximate the
shape of the single-class training data. This is possible
by using kernel functions.\\
\mbox{}\\ 
{\bf Flexible Data Description:}

The Support vector data description is made flexible by
replacing the inner product $ \dotp{x_{i}}{x_{j}} $ with a
suitable kernel function $ K(x_{i},x_{j}) $. The Gaussian
kernel function used in this paper is defined as
\begin{equation}  
K(x, y)= \mathrm{exp}  \dfrac{ -\|x - y\|^2}{2s^2}
 \end{equation}
where $s$ is the Gaussian bandwidth parameter.

The modified mathematical formulation of SVDD with a kernel function is as follows:

Objective:
\begin{equation}  
   \mathrm{max}~ \sum_{i=1}^{n}\alpha _{i}K(x_{i},x_{i}) - \sum_{i,j}^{ }\alpha _{i}\alpha _{j}K(x_{i},x_{j}),
\end{equation}
Subject to:
\begin{align}
\sum_{i=1}^{n}\alpha _{i} &= 1  \\
0 \leq \alpha_{i}\leq C, \forall i=1,\dots,n 
\end{align}
In perfect analogy with the previous section, any $x_i$ with
$\alpha_i = 0$ is an \emph{inside} point, any $x_i$ for
which $\alpha_i > 0$ is called a support vector.

$SV_{<~C}$ is similarly defined as $\{x_j : 0 < \alpha_j < C\}$ 
and the threshold $R^{2}$ is calculated, for any $x_k \in SV_{<~C}$, as
\begin{multline}
R^{2} = K(x_{k},x_{k})-2\sum_{i}^{ }\alpha _{i}K(x_{i},x_{k})+\sum_{i,j}^{ }\alpha _{i}\alpha _{j}K(x_{i},x_{k})
\end{multline}
The value of $R^2$ does not depend on which $x_k \in SV_{<~C}$ is used.
\mbox{}\\
{\bf Scoring:}
For any observation $z$ the distance $\text{dist}^{2}(z)
$ is calculated as follows: 
\begin{equation}
\text{dist}^{2}(z)= K(z,z) - 2\sum_{i}^{ }\alpha_{i}K(x_{i},z)+
\sum_{i,j}^{ }\alpha_{i}\alpha_{j}K(x_{i},x_{j})
\end{equation}
Any point $z$ for which $\text{dist}^{2}(z) > R^{2} $ is designated as an outlier.

\subsection{Importance of the Kernel Bandwidth Value}
In practice, SVDD is almost always computed by using the Gaussian
kernel function, and it is important to set the value of
bandwidth parameter correctly. A small bandwidth leads to
overfitting, and the resulting SVDD classifier overestimates
the number of anomalies. A large bandwidth leads to
underfitting, and many anomalies cannot be detected by the
classifier.

Because SVDD is an unsupervised learning technique, it is
desirable to have an automatic, unsupervised bandwidth selection
technique that does not depend on labeled data that
separate the \emph{inliers} from the \emph{outliers}.
In~\cite{2016arXiv160205257K}, Kakde \etal present the peak criterion,
which is an unsupervised bandwidth selection technique, and show
that it performs better than alternative unsupervised
methods. However, determining the bandwidth suggested
by the peak criterion requires that the SVDD
solution be computed multiple times for the training data for a list
of bandwidth values that lie on a grid. Even though using
sampling techniques can speed up the computation
(see \cite{2016arXiv161100058P}); this method is still expensive.
Moreover to avoid unnecessary computations it is also necessary 
to initiate the grid search at a good starting value and it is
not immediately obvious what a good starting value is. In
this paper we suggest two new criteria : the mean criterion
and the median criterion. The mean criterion has a simple
closed-form expression in terms of the training data.

We evaluated the mean criterion and the median criterion
in multiple ways. We conducted simulation studies where
we could objectively determine the quality of a particular
bandwidth. We compared the results obtained from the mean
and median criteria with those obtained from alternative
methods on a wide range of data sets. The data were specially
selected to probe potential weaknesses in the mean
criterion.

Our results show that the mean criterion is competitive
with the peak and median criteria for most data sets in our
test suite. In addition, computation of the mean criterion 
is fast even when the data set is large. These properties make
the mean criterion a good bandwidth selection technique.
However, unsupervised bandwidth tuning is an extremely
difficult problem, so it is quite possible that there is a
class of data sets for which the mean criterion does not give
good results.

The rest of the paper is organized as follows.
Section~\ref{sec:mcrit} defines the mean and median
criteria for bandwidth tuning, and the remaining sections
compare the mean, median, and peak criteria with each other.

\section{The Mean Criterion for Bandwidth Selection} \label{sec:mcrit}
\subsection{Training Data That Have Distinct Observations}
Assume we have a training data set $x_1,\dots,x_N$ that consists of $N$ distinct
points in $\mathbb{R}^p$ and we want to determine a good kernel bandwidth value
for training this data set.

Given a candidate bandwidth $s$, let $K_N(s)$ denote the $N
\times N$ kernel matrix whose element in position $ij$ is
$\mathrm{exp}(-\dfrac{\|x_i-x_j\|^2}{2s^2}).$ A \emph{tiny}
s is not a good candidate for the kernel bandwidth, because
as $s\rightarrow 0^{+}$, $K_N(s)$ converges to the identity matrix
of order $N$. When the kernel matrix is very close to the
identity matrix, all observations in the original data set
become support vectors. This indicates a case of severe
overfitting.

So for a reasonable bandwidth value $s$, the corresponding
kernel matrix $K_N(s)$ must be sufficiently different from
the identity matrix $I_N$. One way to ensure this would be to choose $s$ such
that
\begin{equation}
\label{eq:1}
\|K_N(s) - I_N\| \geq \delta \|I_N\|
\end{equation}
where $\|\,\|$ is an appropriate matrix norm and $0 < \delta
< 1$ is a tolerance factor. Larger values of $\delta$ will
ensure greater distance from the identity matrix.

It is easy to determine an $s$ that satisfies \eqref{eq:1}
when $\|\,\|$ is chosen as the Frobenius norm. The Frobenius
norm of a matrix is defined as the square root of the sum
of squares of all elements in the matrix; that is, $\|A\| =
\sqrt{\operatorname{trace}(A^TA)}.$

If $\|\,\|$ is the Frobenious norm, then 
\begin{align*}
& \|K_N(s) - I_N\| \geq \delta \|I_N\| \\
\iff & \|K_N(s) - I_N\|^2 \geq \delta^2 N \\
\iff & 2\sum_{i < j} \mathrm{exp}( -\dfrac{\|x_i - x_j\|^2}{s^2} ) \geq \delta^2 N\\
\iff & 2\binom{N}{2}\dfrac{\sum_{i < j} \mathrm{exp}( -\dfrac{\|x_i - x_j\|^2}{s^2} )}{\binom{N}{2}} \geq \delta^2 N
\end{align*}
From the well-known inequality of arithmetic and geometric means, we have
$$
\dfrac{\sum_{i < j} \mathrm{exp}( -\dfrac{\|x_i - x_j\|^2}{s^2} )}{\binom{N}{2}} \geq  \mathrm{exp}\left( -\dfrac{\sum_{i < j}\dfrac{\|x_i - x_j\|^2}{s^2}}{\binom{N}{2}} \right)
$$
so it is sufficient to choose an $s$ such that
\begin{align}
 & 2\binom{N}{2}\exp\left( -\dfrac{\sum_{i < j}\dfrac{\|x_i - x_j\|^2}{s^2}}{\binom{N}{2}} \right) \geq \delta^2 N
\nonumber \\
\iff s &\geq \sqrt{\frac{\bar{D}^2}{\ln\frac{N-1}{\delta^2}}} \label{eq:mdef}
\end{align}
where $\bar{D}^2=\dfrac{\sum_{i<j}||x_i-x_j||^2}{\binom{N}{2}}.$ 

Equation \eqref{eq:mdef} suggests using \begin{equation}s =
\sqrt{\dfrac{\bar{D}^2}{\ln\frac{N-1}{\delta^2}}} \label{eq:mdef1}\end{equation} as the
kernel bandwidth. The numerator of \eqref{eq:mdef} contains
the mean of pairwise squared distances; this suggests creating new
criteria by replacing them with another measure
of central tendency of the squared distances 
in the numerator of \eqref{eq:mdef}. For example we can have
another criterion which suggests using $s =
\sqrt{\dfrac{ \mathrm{median}_{i < j} \|x_i - x_j\|^2 }{\ln\frac{N-1}{\delta^2}}}$ as the bandwidth value.

\begin{tcolorbox}
We call using  $\sqrt{\dfrac{\bar{D}^2}{\ln\frac{N-1}{\delta^2}}}$ as the bandwidth the {\bf mean criterion} for bandwidth
selection, and we call using $\sqrt{\dfrac{ \operatorname{median}_{i < j} \|x_i - x_j\|^2
}{\ln\frac{N-1}{\delta^2}}} = \dfrac{\operatorname{median}_{i < j} \|x_i - x_j\|}{\sqrt{\ln\frac{N-1}{\delta^2}}}$
as the bandwidth the {\bf median criterion} for bandwidth selection.
\end{tcolorbox}

The mean criterion bandwidth can be computed very quickly. 

Let $x_i = \begin{pmatrix} x_{i1} & \dots & x_{ip}
\end{pmatrix}$, let $\mu_1,\dots, \mu_p$ denote the
column means, and let $\sigma_1^2,\dots,\sigma_p^2$ denote
the column variance. So
\begin{align*}
\mu_1 = \dfrac{\sum_{i=1}^{N} x_{i1}}{N} & & \sigma_1^2 = \dfrac{\sum_{i=1}^N( x_{i1} - \mu_1)^2}{N}\\
\mu_2 = \dfrac{\sum_{i=1}^{N} x_{i2}}{N} & &\sigma_2^2 = \dfrac{\sum_{i=1}^N( x_{i2} - \mu_2)^2}{N}\\
... & & ... \\
\mu_p = \dfrac{\sum_{i=1}^{N} x_{ip}}{N} & & \sigma_p^2 = \dfrac{\sum_{i=1}^N( x_{ip} - \mu_p)^2}{N}
\end{align*}
Then it is easy to show that $\bar{D}^2 = \dfrac{2N}{N-1} \sum_{j=1}^p
\sigma_p^2$. So the bandwidth suggested by the mean criterion is
\begin{equation}
\label{eq:meansimple}
s = \sqrt{ \dfrac{2N \sum_{j=1}^p \sigma_p^2}{(N-1) \ln\left(\dfrac{N-1}{\delta^2}\right)}}
\end{equation}
To see this, note the following:
\begin{align*}
\sum_{i < j} \|x_i - x_j\|^2 &= \dfrac{1}{2} \sum_{i=1}^{N}\sum_{j=1}^{N} \|x_i  - x_j\|^2\\
& = \dfrac{1}{2}\sum_{i=1}^{N}\sum_{j=1}^{N} \left(\|x_i\|^2 + \|x_j\|^2 - 2 \langle x_i,x_j \rangle\right)\\
&= \dfrac{1}{2} \left(2N\sum_{i=1}^{N}\|x_i\|^2 - 2\| \sum_{i=1}^N x_i \|^2\right) \\
&= N\sum_{j=1}^{p}\sum_{i=1}^N x_{ij}^2 - \| N \begin{pmatrix} \mu_1 & \dots & \mu_p \end{pmatrix} \|^2\\ 
&= N^2\sum_{j=1}^p (\sigma_j^2 + \mu_j^2) - N^2\sum_{j=1}^p \mu_j^2\\
& = N^2 \sum_{j=1}^p \sigma_j^2
\end{align*}
The result is immediate.

Because the column variances can be calculated in one pass over the data,
the computation of the mean criterion is an $\mathcal{O}(Np)$ algorithm. 

The computation needed for the median criterion cannot be simplified; however,
one can take a sample $x_1^{*},\dots,x_n^{*}$ from the data and use 
$\mathrm{median}_{i < j} \|x_i^{*} - x_j^{*}\|$ as an approximation to 
$\mathrm{median}_{i < j} \|x_i - x_j\|.$ 
\subsection{Training Data That Have Repeated Observations}
We now consider the case where we have repeated observations in the training data set.
Let $\{x_1,\dots,x_N\}$ be the set of distinct points in our training data set in $\mathbb{R}^p$,
and assume that $x_i$ is repeated $w_i > 0$ times. In this case,
the kernel matrix, $K_W(s)$ is a square matrix of
order $W = \sum_{i} w_i$. The kernel matrix can be partitioned into $N^2$ blocks where the
$ij^{\text{th}}$ block is a matrix of order $w_i \times w_j$ given by
$\mathrm{exp}\left( -\dfrac{\|x_i - x_j\|^2}{2s^2} \right) \mathbf{1}_{w_i} \mathbf{1}^{\top}_{w_j}$, where
$\mathbf{1}_{w_k}$ is a column vector of $w_k$ ones. As $s \to 0^+$, $K_W(s)$ converges to $K_W(0)$, a block diagonal matrix with
diagonal blocks $\mathbf{1}_{w_i} \mathbf{1}^{\top}_{w_i}$ for $ 1 \leq i \leq N.$
In this case, we similarly seek an $s$ such that
\begin{equation} 
\label{eq:condw} 
\|K_W(s) - K_W(0)\| \geq \delta \|K_W(0)\| 
\end{equation}
Define:\\
$W = \sum_{i=1}^{N} w_i,\quad \mu = \dfrac{\sum_{i=1}^{N} w_i x_i}{W},\quad M = \sum_{i=1}^{N} w_i^2$,\\
$Q = (W^2 - M)/2  = \sum_{\substack{i < j \\ 1 \leq i,j \leq N}} w_i w_j$, and\\
$\sigma^2 = \left(\sigma_1^2, \dots, \sigma_p^2\right) = \dfrac{ \sum_{i=1}^{N} w_i \|x_i - \mu\|^2}{W}$.

In a manner similar to the previous section, we have
\begin{align*}
 \|K_W(s) - K_W(0)\| & \geq \delta \|K_W(0)\| &\\
\iff  \|K_W(s) - K_W(0)\|^2 & \geq \delta^2 M &\\
\iff  2Q\sum_{\substack{i < j \\ 1 \leq i,j \leq N}} \dfrac{w_i w_j}{Q} \exp( -\dfrac{\|x_i - x_j\|^2}{s^2} ) &\geq
\delta^2 M &\\
\end{align*}
Using Jensen's inequality, 
\begin{align*}
\impliedby 2Q \exp\left( -\sum_{i < j}\dfrac{ w_i w_j\|x_i - x_j\|^2}{Qs^2} \right) & \geq \delta^2 M &\\
\iff s^2 \geq \dfrac{\displaystyle\sum_{\substack{i < j\\ 1 \leq i, j \leq N}} w_i w_j\|x_i - x_j\|^2}{Q
\ln\left(\dfrac{2Q}{\delta^2M}\right)} &
\end{align*}

As in the previous section the preceding bound can be
simplified and expressed in terms of the weighted column
variances. We have any $s$ that satisfies the following inequality also satisfies \eqref{eq:condw}.
\begin{equation}
\label{eq:meanw}
s \geq \sqrt{\dfrac{W^2\sum_{i=1}^{p} \sigma_i^2}{Q \ln\left(\dfrac{2Q}{\delta^2M}\right)}}
\end{equation}

As expected, \eqref{eq:meansimple} equals \eqref{eq:meanw} when $w_1 = w_2 = \dots = w_N$.
Equation \eqref{eq:meanw} is derived for completeness; it will not be used in the rest of this
article. We will use \eqref{eq:mdef1} throughout this article.

\subsection{Choice of $\delta$}
The mean and median criteria depend on the parameter $\delta$. For the mean and median criteria
to be effective, there should be an easy way to choose the value
$\delta$. Otherwise we will have simply replaced the difficult problem of
choosing a bandwidth with another difficult problem of choosing $\delta$. In
our investigations, we noticed that setting $\delta$ to $\sqrt{2} \times 10^{-6}$ works for
most cases. So unless explicitly stated otherwise, the value of $\delta$ is
$\sqrt{2} \times 10^{-6}$ throughout this article.

\section{Evaluating the Mean criterion}
\subsection{Alternative Criteria}
In \citep{Aggarwal:2013:OA:2436823}, Aggarwal suggests setting
$\sigma=\mathrm{median}_{i < j} \|x_i - x_j\|$ for the kernel that is
parametrized as $\mathrm{exp}\left(-\dfrac{\|x-y\|^2}{\sigma^2}\right)$, which
translates to using $s=\sigma/\sqrt{2}$ for the kernel parametrized as
$\mathrm{exp}\left(-\dfrac{\|x-y\|^2}{2s^2}\right)$. We call this the \medto~criterion.
We compare the mean and median criteria with the \medto~criterion.
In addition, we compare the mean and median criterion with the peak criterion (see ~\citep{2016arXiv160205257K}).
Since the peak criterion performs better than the
alternative criteria mentioned in \citep{2016arXiv160205257K}, we omit
the other criteria that are mentioned there.

\subsection{Choice of Data sets}
The mean and median criteria depend on the distribution of
pairwise distances of the training data set. These methods might not
work well if the distribution of the pairwise distances is
skewed. So it is important to check the performance of the
mean and median criteria on data sets that have a skewed
distribution of pairwise distances.

The distribution of pairwise distances in data sets where the data lie in
distinct clusters is typically multimodal and skewed. See Figure~\ref{fig:3cscatter} for 
a data set that has three distinct clusters; the histogram of pairwise distances as seen in
Figure~\ref{fig:3chist} indicates a skewed and multimodal distribution.

For this reason, we check the performance of the different criteria on
``connected'' data that is, data without any clusters and on data sets where there
are two or more clusters.

\section{Comparing the Criteria on Two-Dimensional Data}

\subsection{Two-Dimensional Connected Data}
\subsubsection{Data Description}
\begin{figure}
\caption{Results for banana data. The darkly shaded region is the inlier region.}
\label{fig:banana}
\subfigure[Scatter plot of banana-shaped data]{\label{fig:bananascatter}
\includegraphics[width=\mywidth]{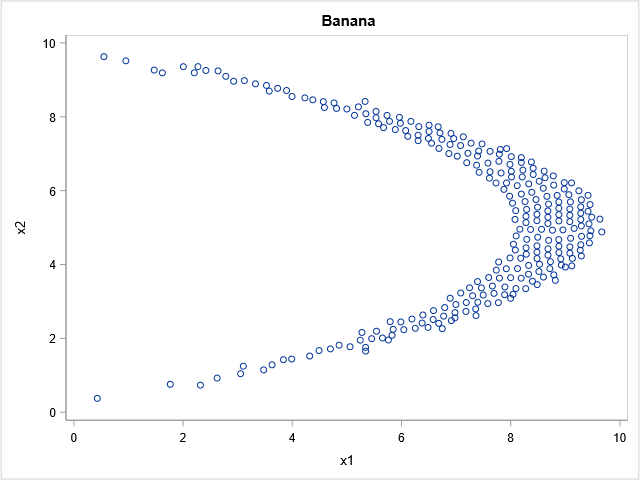}}
\subfigure[Mean criterion result]{\label{fig:bananamean}\includegraphics[width=\mywidth]{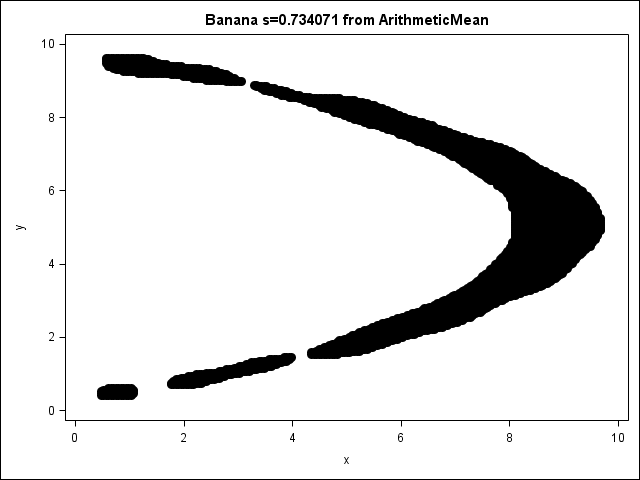}}
\subfigure[Median criterion result]{\label{fig:bananamedian}\includegraphics[width=\mywidth]{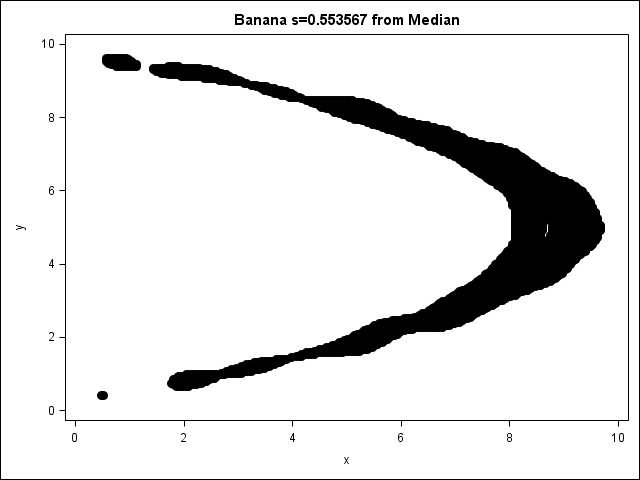}}
\subfigure[Peak criterion result]{\label{fig:bananapeak}\includegraphics[width=\mywidth]{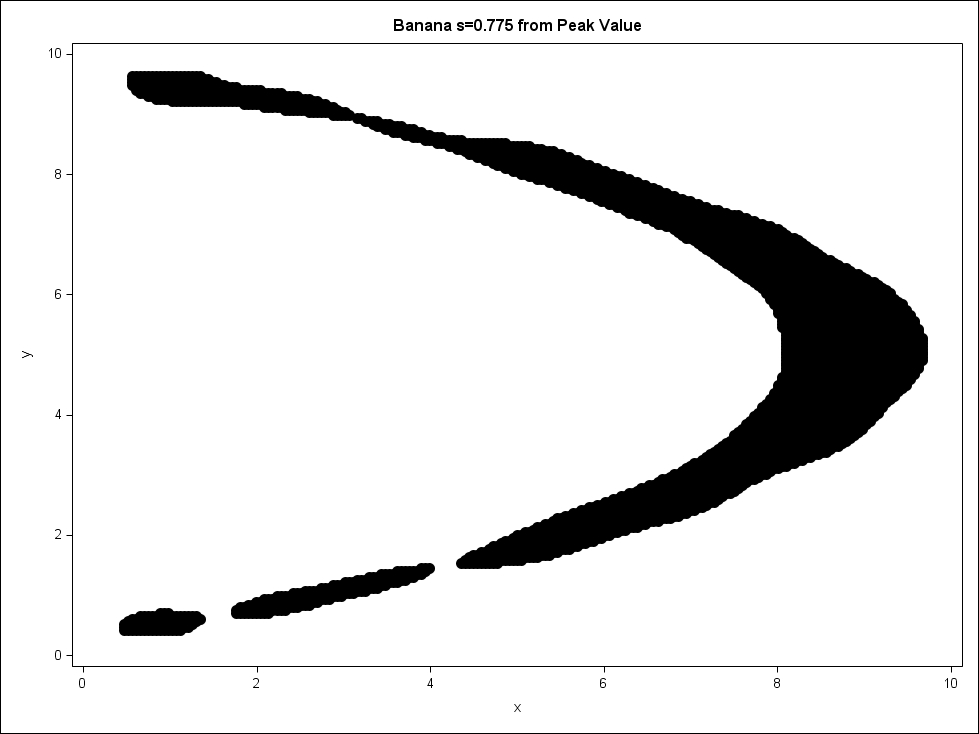}}
\subfigure[\Medto~criterion result]{\label{fig:bananamedian2}\includegraphics[width=\mywidth]{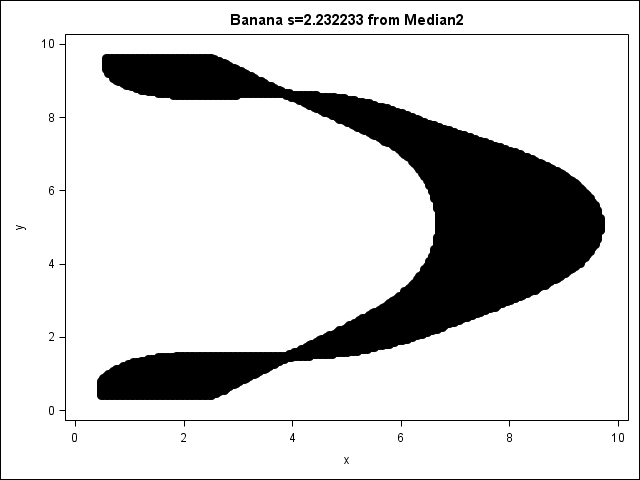}}
\end{figure}

\begin{figure}
\caption{Results for star data. The darkly shaded region is the inlier region.}
\label{fig:star}
\subfigure[Scatter plot of star data]{\label{fig:starscatter}\includegraphics[width=\mywidth]{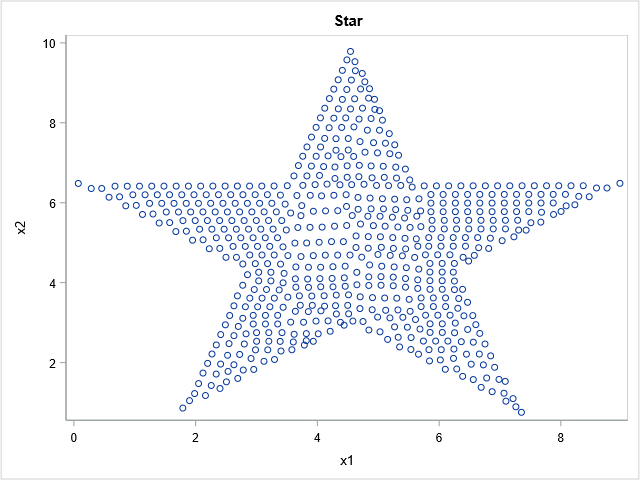}}
\subfigure[Mean criterion result]{\label{fig:starmean}\includegraphics[width=\mywidth]{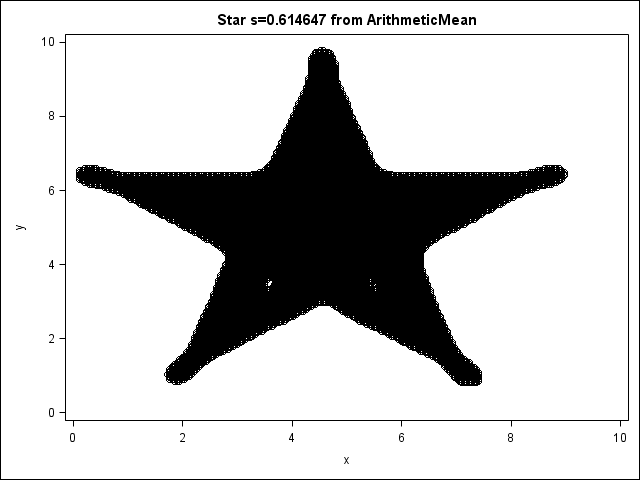}}
\subfigure[Median criterion result]{\label{fig:bananamedian}\includegraphics[width=\mywidth]{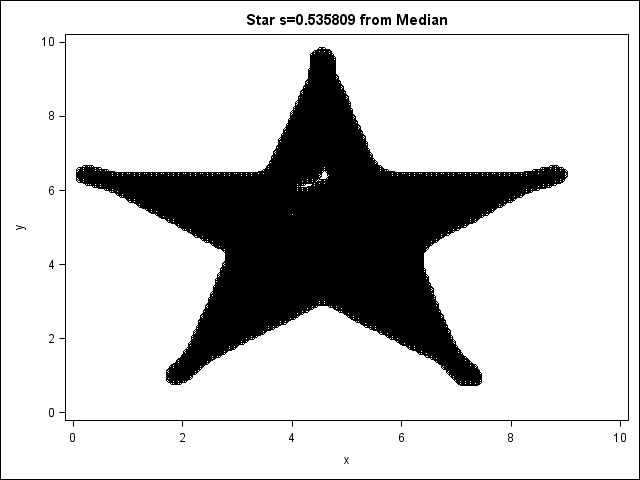}}
\subfigure[Peak criterion result]{\label{fig:bananapeak}\includegraphics[width=\mywidth]{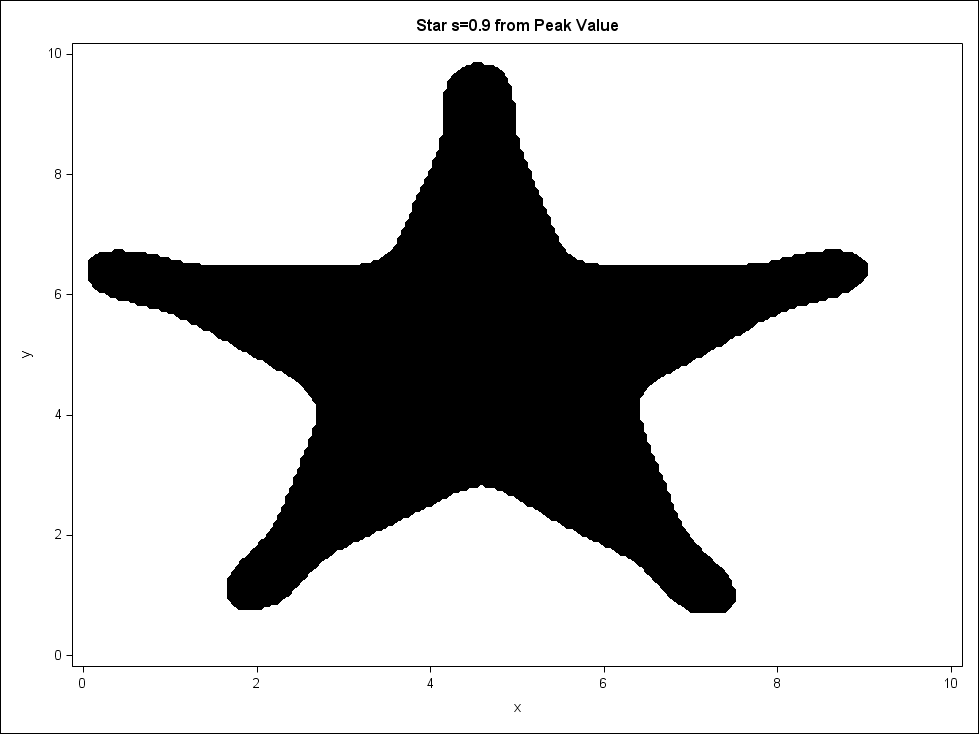}}
\subfigure[\Medto~criterion result]{
\label{fig:bananamedian3}
\includegraphics[width=\mywidth]{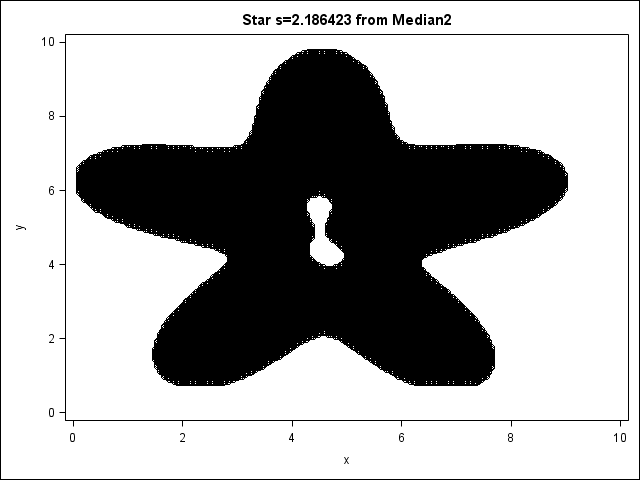}
}
\end{figure}

In this section, we compare the performance of the mean, median, \medto, and peak
criteria on selected two-dimensional data. These data sets are connected; that is,
there are no clusters in the data. Because the data are two-dimensional, we can
visually evaluate the quality of results. To evaluate the results, we obtain the
data description provided by the different bandwidths, and then we score the
bounding rectangle of the data by dividing it into a $200 \times 200$ grid. The
inlier region that is obtained from scoring should closely match the training data.
Figure \ref{fig:banana} displays the results for a banana-shaped data, and Figure
\ref{fig:star} displays the results for a star-shaped data.

\subsubsection{Conclusion}
The scoring results indicate that the bandwidth values
computed using the mean and median criteria provide a data
description of good quality. Such a description is close to
the one obtained using the peak criteria.
The \medto~criterion does not work well for these data sets.

\subsection{Two-Dimensional Disconnected Data}
\subsubsection{Data Description}
\begin{figure}
\caption{Results for the three-clusters data. The darkly shaded region is the inlier region.}
\label{fig:3c}
\subfigure[Scatter plot of three-clusters data]{\label{fig:3cscatter}
\includegraphics[width=\mywidth]{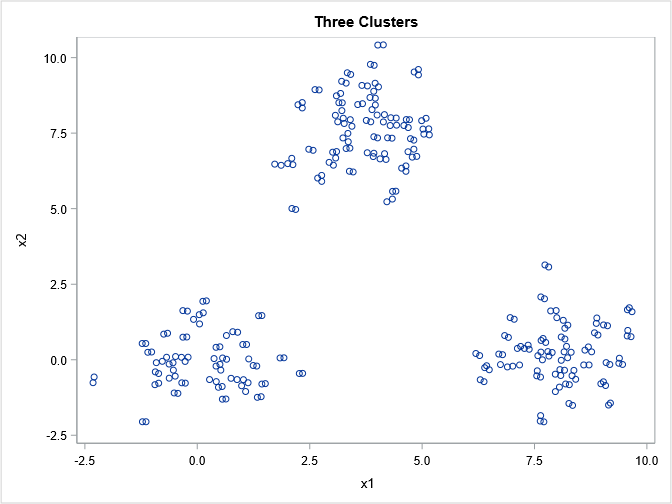}}
\subfigure[Histogram of pairwise distances]{\label{fig:3chist}\includegraphics[width=\mywidth]{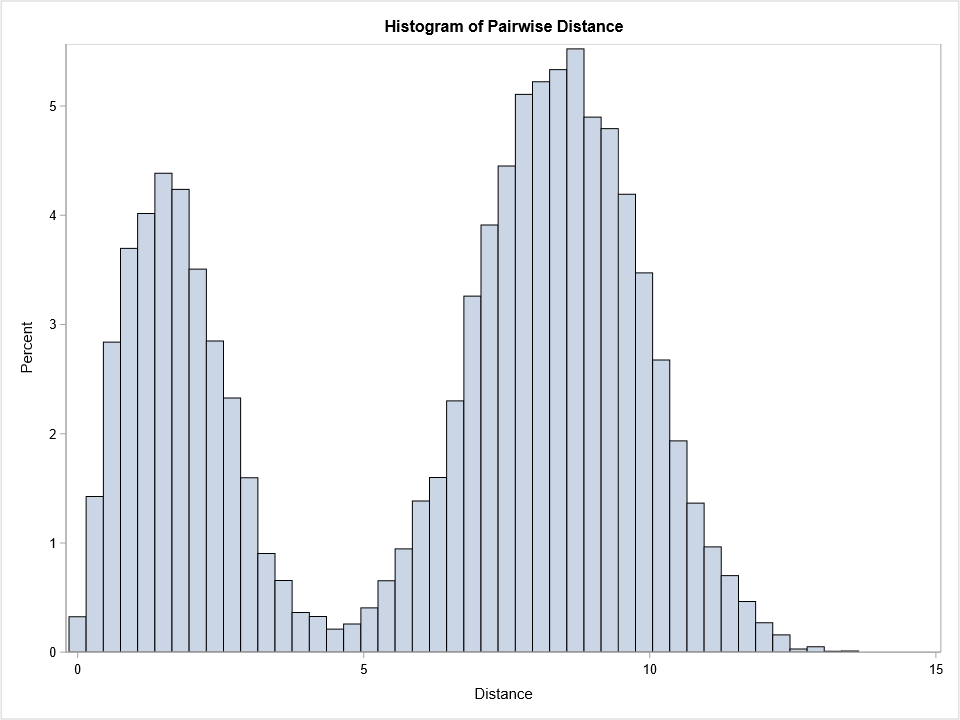}}
\subfigure[mean criterion result]{\label{fig:3cmean}\includegraphics[width=\mywidth]{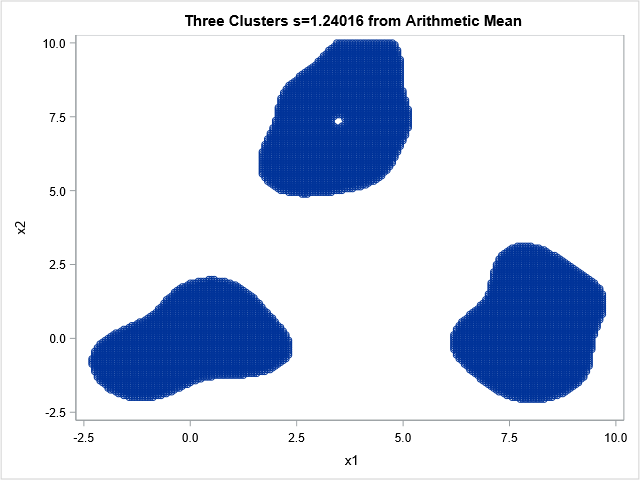}}
\subfigure[median criterion
result]{\label{fig:3cmedian}\includegraphics[width=\mywidth]{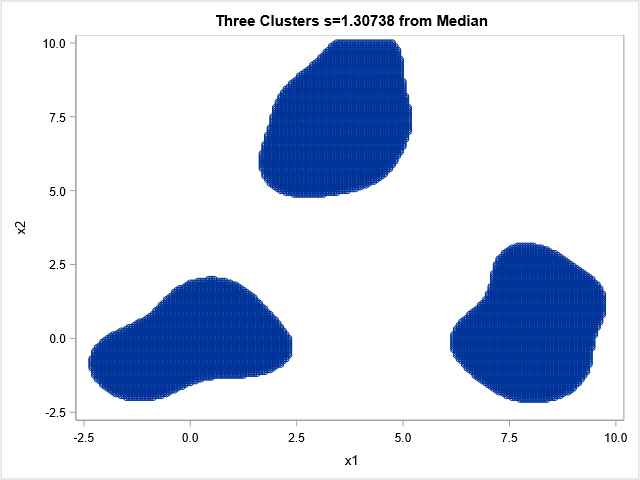}}\quad\quad\   
\subfigure[Peak criterion result]{\label{fig:3cpeak}\includegraphics[width=\mywidth]{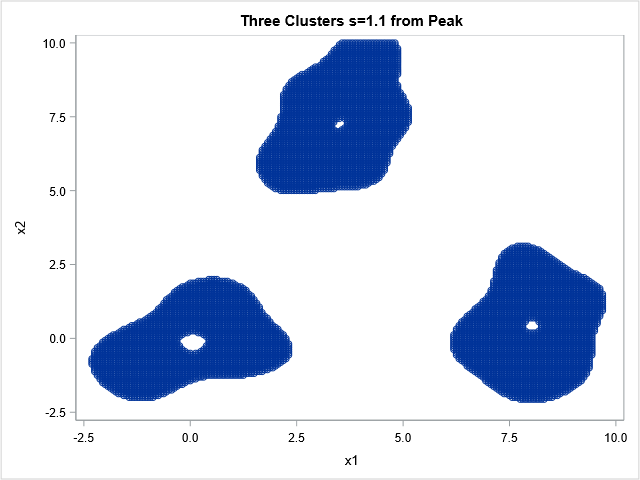}}\hfill%
\subfigure[\medto~criterion result]{
\label{fig:3cm2}
\includegraphics[width=\mywidth]{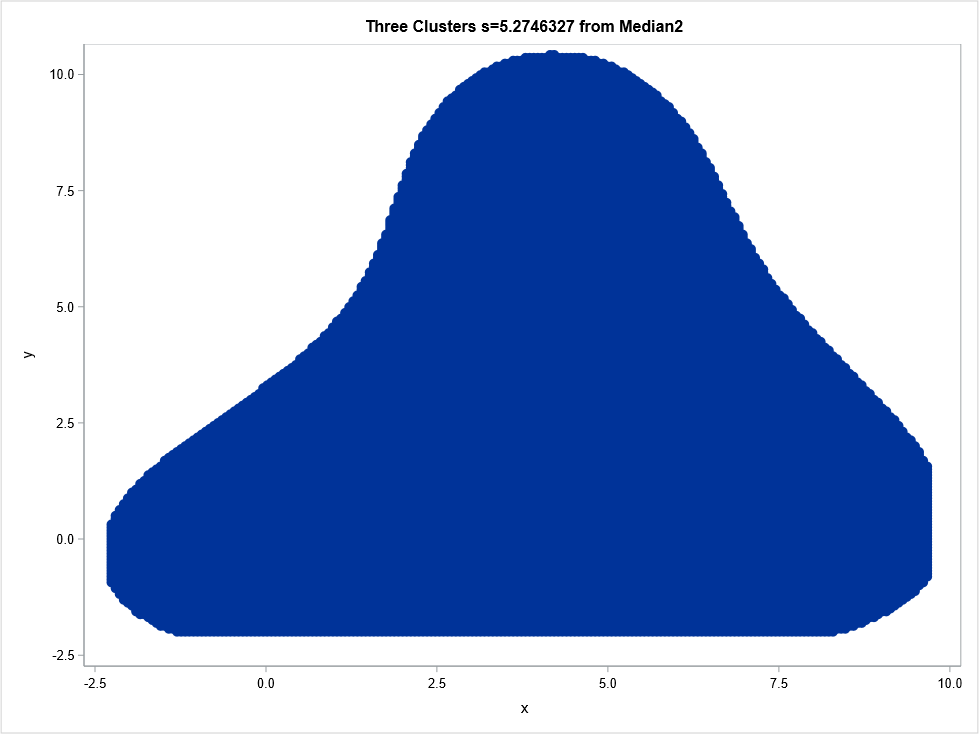}
}
\end{figure}

\begin{figure}
\caption{Results for the refrigerant data. The darkly shaded region is the inlier region.}
\label{fig:f}
\subfigure[Scatterplot]{\label{fig:rf} \includegraphics[width=\mywidth]{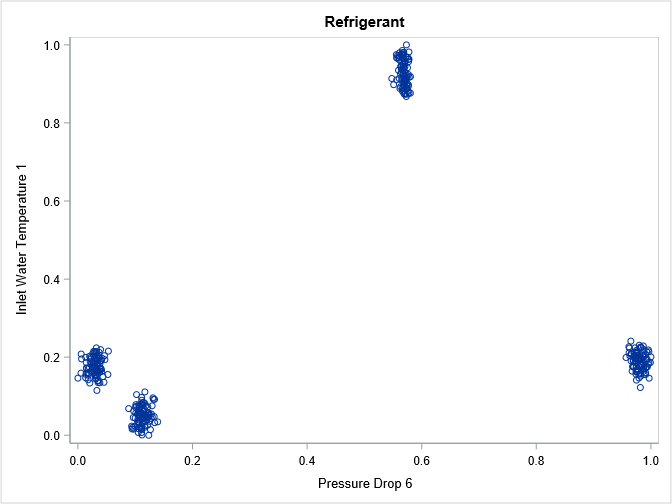}}
\subfigure[Histogram of pairwise distances]{\label{fig:rfh} \includegraphics[width=\mywidth]{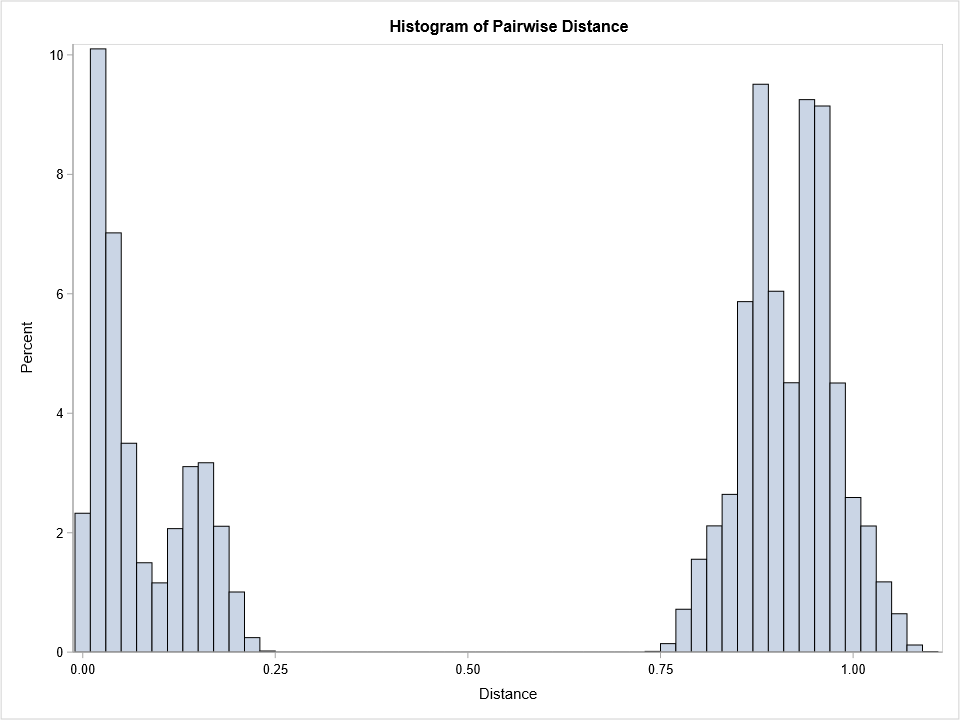}}
\subfigure[Mean criterion result]{\label{fig:fmean}\includegraphics[width=\mywidth]{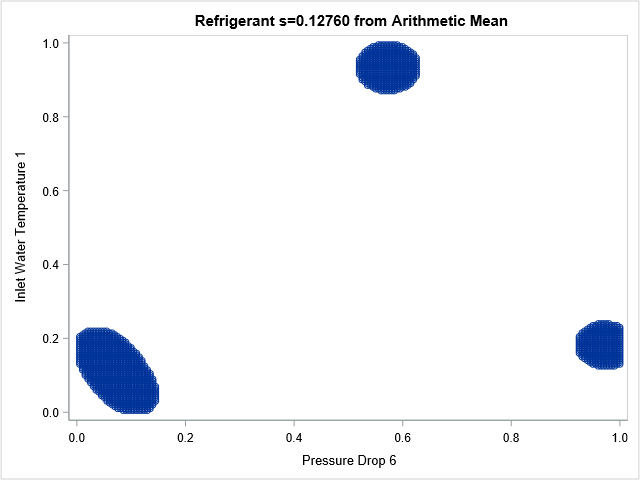}}
\subfigure[Median criterion result]{\label{fig:fmedian}\includegraphics[width=\mywidth]{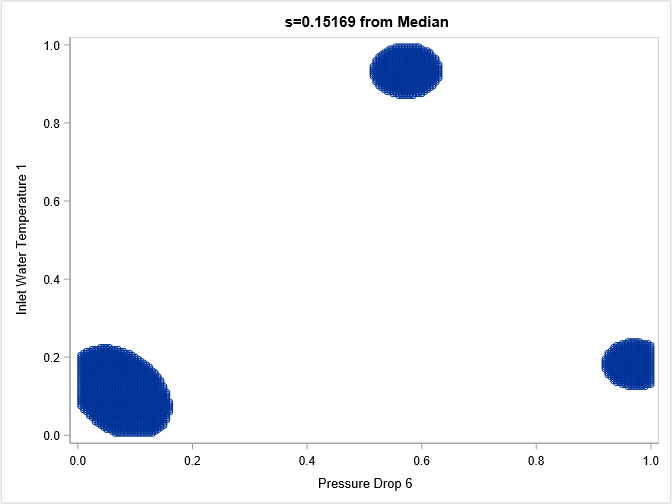}}
\subfigure[Peak criterion result]{\label{fig:fpeak}\includegraphics[width=\mywidth]{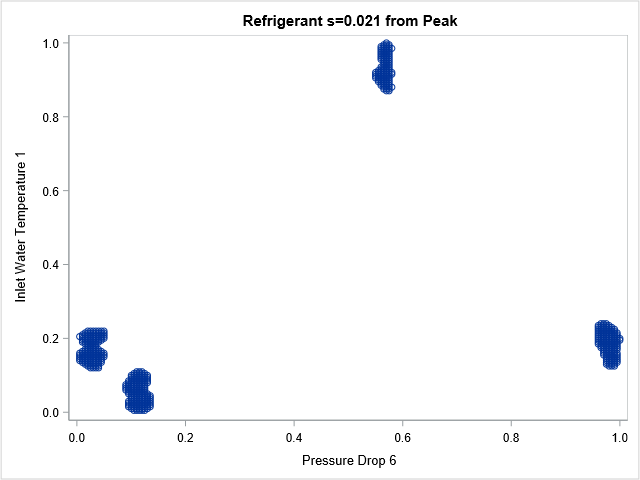}}\hfill%
\subfigure[\Medto~criterion result]{\label{fig:fm2}\includegraphics[width=\mywidth]{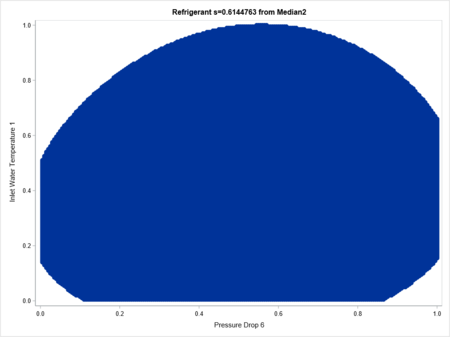}}
\end{figure}

\begin{figure}
\caption{Results for the two-donuts and munchkin data. The darkly shaded region is the inlier region.}
\label{fig:2dm}
\subfigure[Scatter plot]{\label{fig:2dmscatter}
\includegraphics[width=\mywidth]{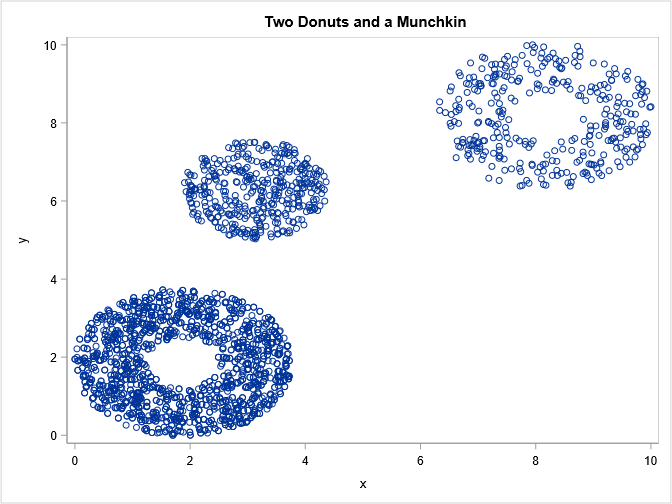}}
\subfigure[Histogram of pairwise distances]{\label{fig:2dmhist}
\includegraphics[width=\mywidth]{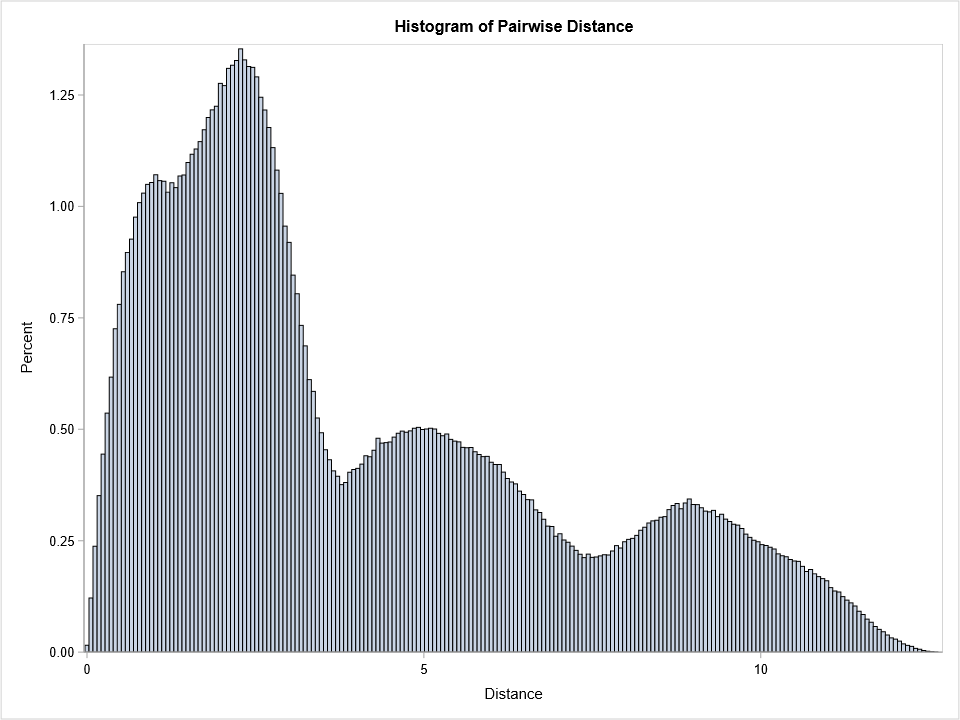}}
\subfigure[Mean criterion result]{\label{fig:2dmmean}\includegraphics[width=\mywidth]{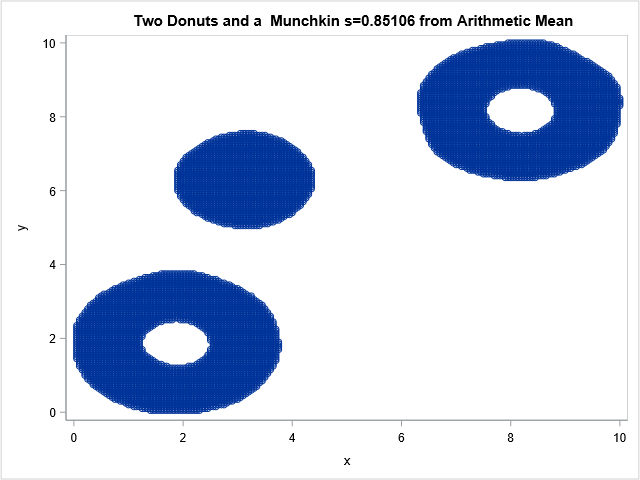}}
\subfigure[median criterion result]{\label{fig:2dmmedian}\includegraphics[width=\mywidth]{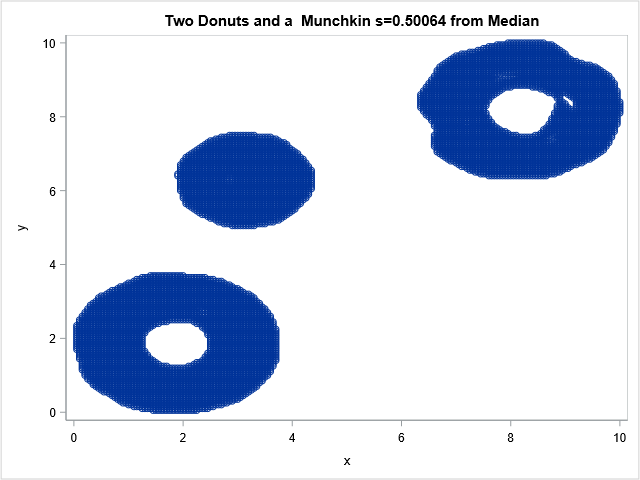}}
\subfigure[Peak criterion result]{\label{fig:2dmpeak}\includegraphics[width=\mywidth]{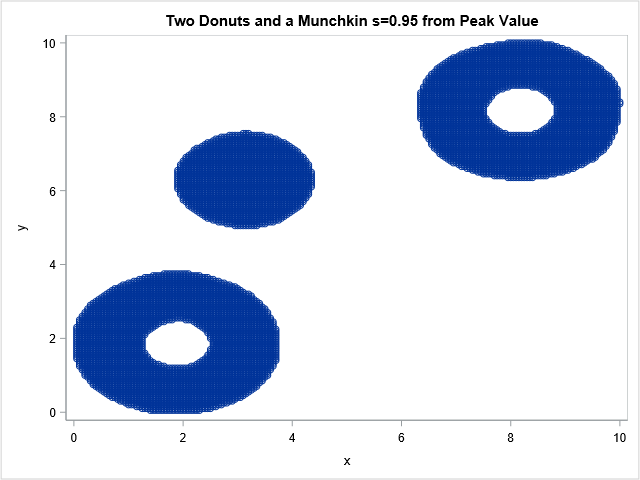}}\hfill%
\subfigure[\medto~criterion result]{\label{fig:2dm2}\includegraphics[width=\mywidth]{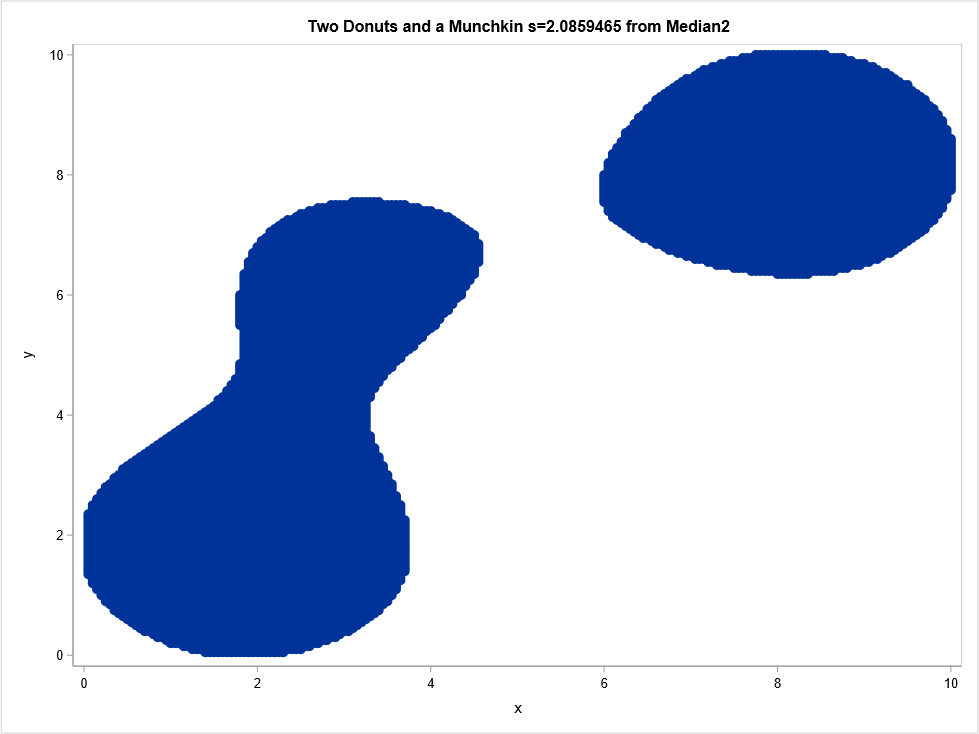}}
\end{figure}

In this section, we compare the performance of the mean, median, \medto, and peak criteria on 
selected two-dimensional data that lie in different clusters.
Selecting the bandwidth of such data is usually more difficult than estimating the bandwidth of
connected data.
Because the data are two-dimensional, we can visually evaluate the quality of results.
To evaluate the results, we obtain the data description that is provided by the different
bandwidths, and then we score the bounding rectangle of the data by dividing it
into a $200 \times 200$ grid. The inlier region obtained from scoring should
closely match the training data.
The following data sets are used in this section: 
\begin{enumerate}
\item The three-clusters data, which consists of three clusters \cite{institute2015sas}.
Figure \ref{fig:3c} displays the data and the scoring results.
\item The refrigerant data, which consists of four clusters \cite{heck:2000}.
Figure \ref{fig:f} displays the data and the scoring results.
\item A simulated ``two-donuts and a munchkin" data set which consists of two donut-shaped regions
and a spherical region.
Figure \ref{fig:2dm} displays the data and the scoring results.
\end{enumerate}


\subsubsection{Conclusion}
The scoring results indicate that the bandwidth value that is computed using the mean
and median criteria provides a data description of reasonably good quality for the
three-cluster data set and for the two-donuts and munchkin data set. Such description is close to
the one obtained using the peak criterion. 

For the refrigerant data set, the peak criterion significantly outperforms other
methods. The data description obtained from the peak criterion can separate
out all four clusters, whereas the other methods merge two clusters that lie
close to each other. Although the mean and median criterion do not perform as
well as the peak criteria, any point in the \emph{inlier} region is close to
the training data, and the area of the region that is misclassified is small
compared to the bounding region of the training data. So the result is still very
reasonable. 

The \medto~criterion again performs poorly on all these data sets, and it is not
considered as an candidate in the remaining sections.

\section{Comparing the Criteria on High-Dimensional Data}
\begin{tcolorbox}
The \fone~score is a common measure of a binary classifier's
accuracy. It is defined as $$\dfrac{2TP}{2TP+FN+FP}$$ where
$TP,FN,$ and $,FP$ stand for the number of true-positives, number of
false-negatives, and number of false-positives, respectively.
\end{tcolorbox}
Comparing the different criteria for high-dimensional
data is much more difficult than comparing them for two-dimensional data. In two-dimensional data,
the quality of the result can be easily judged by looking at the plot of
the scoring results. But this is not possible for high-dimensional data.
For the purpose of evaluation, we selected
labeled high-dimensional data that have a dominant class. We
used SVDD on a subset of the dominant class to obtain a
description of the dominant class, and then we scored the
rest of the data to evaluate the criteria. We expect the
points in the scoring data set that correspond to the dominant
class to be classified as \emph{inliers} and all other points to
be classified as outliers. Because the data are labeled, we
can also use cross validation to determine the bandwidth
that best describes the dominant class in the sense of
maximizing a measure of fit, such as the \fone~score. So in
this section we compare the bandwidth suggested by the
different unsupervised criteria with the bandwidth obtained
through cross validation for various benchmark data sets. The
results are summarized in Table \ref{table:hd} below. The
benchmark data sets used for the analysis are described in
sections~\ref{sub:metal} through \ref{sub:intrusion}.

\begin{\mytable}[ht]
\caption{Results for High-Dimensional Data}
\label{table:hd}
\centering
\begin{tabular}{|p{20mm}|c|c|c|c|c|}
\hline
Data & Dimension/Nobs & Max (\fone)(s) & Peak (\fone)(s) & Mean (\fone)(s) & Median (\fone)(s)\\
\hline 
Metal Etch & 20/96 & (0.8)(0.69) & (0.56)(0.46) & (0.42)(0.24) & (0.43)(0.19) \\
\hline
Shuttle & 9/2000 & (0.96)(17) & (0.96)(14) & (0.95)(11) & (0.84)(5.75) \\
\hline
Spam & 57/1500 & (0.63)(50) & (0.63)(65) & (0.42)(0.24) & (0.43)(0.19) \\
\hline
Tennessee Eastman& 41/2000 & (0.19)(17) & (0.16)(8) & (0.14)(7.22) & (0.135)(6.04)\\
\hline
Intrusion& 45/11490 & (0.95)(7050) & (0.89)(5060) & (0.95)(9667) & (0.95)(356)\\
\hline
\end{tabular}
\end{\mytable}
In Table \ref{table:hd}, the second column contains
the values of the cross validation bandwidth and its
corresponding \fone~score, and the third, fourth and fifth
columns contain the bandwidth suggested by the peak, mean
and median criteria and their corresponding \fone~scores.
\begin{tcolorbox}
{\bf Caveat}\\
SVDD is a geometric classifier, so using labels in this manner is useful
only if they geometrically separate the data. If the labels actually separate
the data geometrically the bandwidth obtained from cross validation will lead
to a high \fone~score.
\end{tcolorbox}

We now describe the data sets mentioned in Table \ref{table:hd}. 
\subsection{Metal Etch Data}
This data set consists of 20 process variables, an ID variable, and a timestamp
variable from a metal wafer etcher. The data consist of measurements from 108 normal wafers
and 21 faulty wafers. The training data set contains half of all the normal wafers, and
the scoring data set contains the remaining observations.\label{sub:metal}
The data set used in this analysis is explained in \cite{wise1999comparison} and
can be obtained from \cite{data:metal}.

\subsection{Shuttle Data}
\label{sub:shuttle}
This data set consists of measurements made on a shuttle.
The data set contains nine numeric attributes and one class attribute. Out of
58,000 total observations, 80\% of the observations belong to class one. A random
sample of 2,000 observations belonging to class 1 was selected for training, and
the remaining 56,000 observations were used for scoring.
This data set is from the UC Irvine Machine Learning Repository \citep{Lichman:2013}. 

\subsection{Spam Data}
\label{sub:spam}
The spam data set consists of emails that were
classified as spam or not. Each record corresponds to an individual email. The
total number of attributes is 57. Most attributes are frequencies of specific
words. Training is performed using a subset of non-spam observations. Remaining
observations, which relate to both the spam and non-spam emails, were used for
scoring. This data set is from the UC Irvine Machine Learning Repository \citep{Lichman:2013}. 


\subsection{Tennessee Eastman}
\label{sub:te}
The data set was generated using the MATLAB simulation code, which provides a
model of an industrial chemical process. The data were
generated for normal operations of the process and twenty faulty processes. Each
observation consists of 41 variables, out of which 22 were measured continuously
every 6 seconds on average and the remaining 19 were sampled at a specified interval
of every 0.1 or 0.25 hours. From the simulated data, we created an analysis
data set that uses the normal operations data of the first 90 minutes and data
that correspond to faults 1 through 20. A data set that contains observations of
normal operations was used for training. Scoring was performed to determine
whether the model could accurately classify an observation as belonging to normal
operation of the process. The MATLAB simulation code is available at \cite{Ricker:2002}.

\subsection{Intrusion Data}
\label{sub:intrusion}
This data set contains multivariate data that characterize cyber attacks. It
contains 45 attributes which include type of service, number of source bytes, number of failed logins, and
number of files created. Out of the 24,156 observations, 22,981 were labeled
no attack and 1,175 were labeled attack. Half of the no attack observations
were used as training data and the remaining observations were used as scoring
data. This data set can be obtained from \cite{kdd:1999}.



\subsection{Conclusion}
The bandwidths suggested by the mean criterion are similar
to the bandwidth suggested by the median and peak criteria
for many data sets. This similarity makes the mean criterion an
attractive bandwidth selection criterion because it can be
computed very quickly.

\section{Simulation Study on Random Polygons}
\subsection{Design}
In this section, we conduct a simulation study to compare the different
bandwidth selection methods. The simulation study consists of training SVDD on 
randomly generated polygons. Given the number
of vertices $N$, we first generate the vertices of the polygon in
counterclockwise direction as $r_1 e^{ \mathbf{i} \theta_{(1)}},r_2
e^{\mathbf{i} \theta_{(2)}},\dots,r_N e^{\mathbf{i}\theta_{(N)}},$ where
$\theta_{(1)},\dots,\theta_{(N)}$ are the order statistics of a uniform iid
sample from the interval $(0,2\pi)$ and $r_1,r_2,\dots, r_N$ are uniformly chosen
from an interval $[r_{\text{min}},r_{\text{max}}]$.

\begin{figure}
\caption{Simulation study with random polygons}
\label{fig:polysim}
\subfigure[True polygon]{\label{fig:polysim:true}
\includegraphics[width=\mywidth]{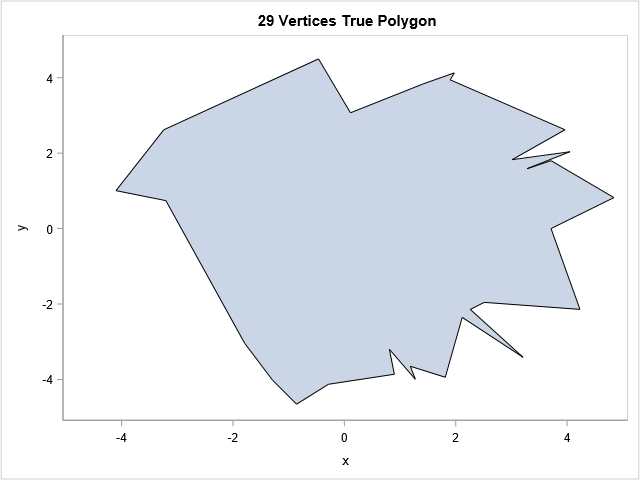}}
\subfigure[Sampled data]{\label{fig:polysim:data}\includegraphics[width=\mywidth]{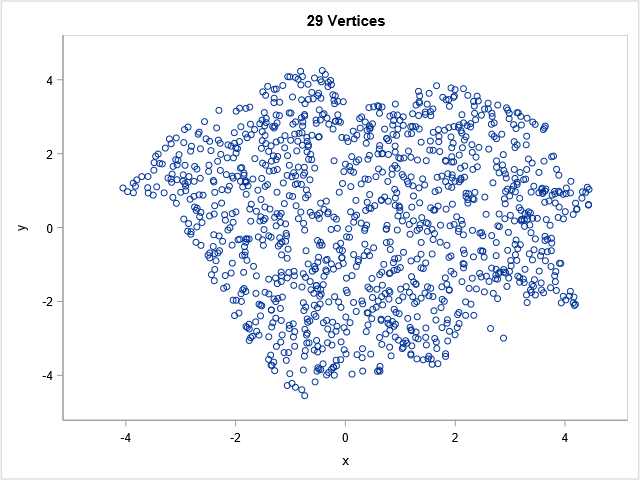}}
\end{figure}

We then sample uniformly from the interior of the polygon and compute the
bandwidths that are suggested by the mean, median, and peak criteria. Because it is easy
to determine whether a point actually lies inside a particular polygon, we can also use
cross validation to determine the best bandwidth parameter. To do so we divide
the bounding rectangle of this polygon into a $200 \times 200$ grid and label each
point in the grid as an inside or outside point depending on whether that
point is inside or outside the polygon. We can choose the best bandwidth as
the one that classifies the grid points as inside or outside points with the
highest \fone~score. Figure~\ref{fig:polysim} shows a typical polygon and 
the data that are generated from the polygon for fitting SVDD.

In our simulation study, we set $r_{\text{min}}=3$ and $r_{\text{max}}=5$ and we
generate polygons whose number of vertices vary from 5 to 30. For a particular
vertex size, we generate 20 polygons. For each such polygon, we create a data set
that consists of 600 points sampled from the interior of the polygon, and we use
this sample to obtain bandwidths that are obtained through cross validation and from the
mean and median criteria. For each such polygon, we have the \fone~score
that corresponds to the cross-validation bandwidth $F_{\text{max}}$ and the
\fone~scores that correspond to the mean and median criterion, $F_{\text{mean}}$
and $F_{\text{median}}$, respectively. $F_{\text{max}}$ is the best possible \fone~score that
can be attained by any bandwidth. At the end of the simulation we have a collection of  ``\fone~score ratios'': $F_{\text{mean}}/F_{\text{max}}$ and
$F_{\text{median}}/F_{\text{max}}$, one for each polygon used in the simulation. If
most of these values are close to $1$, this will indicate
that the bandwidth suggested by the mean and median
criterion is competitive with the bandwidth that maximizes
the \fone~score.

\subsection{Results}
\begin{figure}
\caption{Simulation study results}
\label{fig:box}
\subfigure[Mean criterion]{\label{fig:box:sim:mean}
\includegraphics[width=\mywidth]{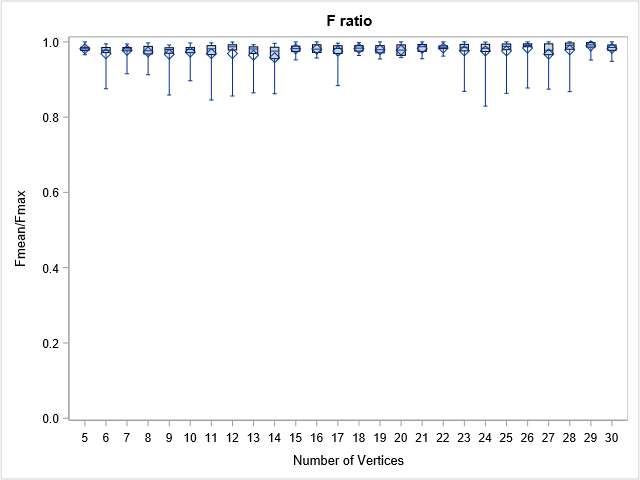}}
\subfigure[Median criterion]{\label{fig:box:sim:median}\includegraphics[width=\mywidth]{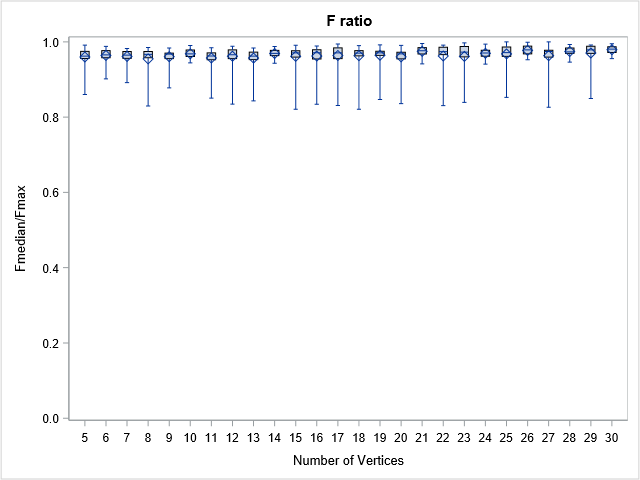}}
\end{figure}

The box-and-whiskers plots in Figure~\ref{fig:box} summarize
the simulation study results. The X axis shows the number
of vertices of the polygon, and the box on the Y axis shows
the distribution of the \fone~scores. The bottom and the top
of the box show the first and third quartile values.
The ends of the whiskers represent the minimum and 
maximum values of the \fone~score ratio. The diamond shape
indicates the mean value, and the horizontal line in the box
indicates the second quartile. The plots shows that the ratio is
greater than 0.8 across all values of number of vertices.
The \fone~score ratio in the top three quartiles is greater
than 0.9 across all values of the number of vertices. As the
complexity of the polygon increases with increasing number
of vertices, the spread of \fone~score ratio also increases.

\subsection{Conclusion}
 The fact that the
\fone~score ratios are always close to 1 suggests the mean and median criteria
generalize across different training data sets. However, a similar simulation
performed for the peak criterion in~\cite{2016arXiv160205257K} shows that the
distribution of \fone~score ratios for the peak criterion is even more concentrated
around 1 that and the minimum values of the \fone~score ratios are much higher that
those for the mean and median criteria. This shows that the peak criterion
generalizes better than the mean and median criterion.

\section{Conclusion and Future Work}
We proposed two new bandwidth selection criteria, the mean
criterion and the median criterion, for the Gaussian kernel for SVDD
training. The proposed criteria give results that are similar to
the peak criterion for many data sets, and hence are a good
starting point for determining a suitable bandwidth for a
particular data set. The suggested criteria might not be the most
appropriate for data sets where the distance matrix is highly
skewed, and more research is needed for determining an
appropriate bandwidth for such cases.

\bibliography{svdd_bw}
\bibliographystyle{plain}
\end{document}